%% file: main.tex
\documentclass[10pt,twocolumn,letterpaper]{article}

\usepackage{cvpr}              


\definecolor{cvprblue}{rgb}{0.21,0.49,0.74}
\usepackage[pagebackref,breaklinks,colorlinks,allcolors=cvprblue]{hyperref}

\usepackage{graphicx}
\usepackage{amsmath}
\usepackage{amssymb}
\usepackage{booktabs}
\usepackage{multirow}
\usepackage{url}
\usepackage{makecell}
\usepackage{algorithmic}
\usepackage{xcolor,colortbl}
\usepackage{graphicx} 
\usepackage{enumitem}
\usepackage{amsmath}
\usepackage{bm}     
\usepackage[normalem]{ulem}
\usepackage{cancel}

\definecolor{cvprblue}{rgb}{0.21,0.49,0.74}
\usepackage[pagebackref,breaklinks,colorlinks,allcolors=cvprblue]{hyperref}

\title{SAISA: Towards Multimodal Large Language Models \\ with Both Training and Inference Efficiency}

\author{
  \textbf{Qianhao Yuan}${}^{1,2,}$\thanks{~These authors contributed equally}, 
  \textbf{Yanjiang Liu}${}^{1,2,*}$, 
  \textbf{Yaojie Lu}${}^{1}$, 
  \textbf{Hongyu Lin}${}^{1}$,
  \textbf{Ben He${}^{1,2}$},
  \textbf{Xianpei Han${}^{1}$},
  \textbf{Le Sun${}^{1}$}
  \\
  ${}^{1}$Chinese Information Processing Laboratory, Institute of Software, Chinese Academy of Sciences\\
  ${}^{2}$University of Chinese Academy of Sciences \\
 \texttt{\{yuanqianhao2024,liuyanjiang2021,luyaojie,hongyu\}@iscas.ac.cn}\\
  \texttt{benhe@ucas.ac.cn}, \texttt{\{xianpei,sunle\}@iscas.ac.cn}
}

\begin{document}
\maketitle
\input{sec/0_abstract}    
\input{sec/1_intro}
\input{sec/2_current}
\input{sec/3_naavit}
\input{sec/4_method}
\input{sec/5_experiments}
\input{sec/6_related_work}
\input{sec/7_conclusion}
{
    \small
    \bibliographystyle{ieeenat_fullname}
    \bibliography{main}
}

\end{document}

%% file: sec/0_abstract.tex
\begin{abstract}

Multimodal Large Language Models (MLLMs) mainly fall into two architectures, each involving a trade-off between training and inference efficiency: embedding space alignment (e.g., LLaVA-1.5) is inefficient during inference, while cross-attention space alignment (e.g., Flamingo) is inefficient in training.
 this paper, we compare these two architectures and identify the key factors for building efficient MLLMs.
A primary difference between them lies in how attention is applied to visual tokens, particularly in their interactions with each other.
To investigate whether attention among visual tokens is necessary, we propose a new self-attention mechanism, NAAViT (\textbf{N}o \textbf{A}ttention \textbf{A}mong \textbf{Vi}sual \textbf{T}okens), h eliminates this type of attention.
Our pilot experiment on LLaVA-1.5 shows that attention among visual tokens is highly redundant.
, we introduce SAISA (\textbf{S}elf-\textbf{A}ttention \textbf{I}nput \textbf{S}pace \textbf{A}lignment), a novel architecture that enhance both training and inference efficiency.
SAISA directly aligns visual features with the input spaces of NAAViT self-attention blocks, reducing computational overhead in both self-attention blocks and feed-forward networks (FFNs).
Using the same configuration as LLaVA-1.5, SAISA reduces inference FLOPs by 66\% and training budget by 26\%, while achieving superior performance in terms of accuracy.
Comprehensive ablation studies further validate the effectiveness of SAISA across various LLMs and visual encoders.
The code and model will be publicly available at \href{https://github.com/icip-cas/SAISA}{https://github.com/icip-cas/SAISA}.
\end{abstract}

%% file: sec/1_intro.tex
\section{Introduction}
\label{sec:intro}

\begin{figure}[t!]
\centering
\includegraphics[width=\linewidth]{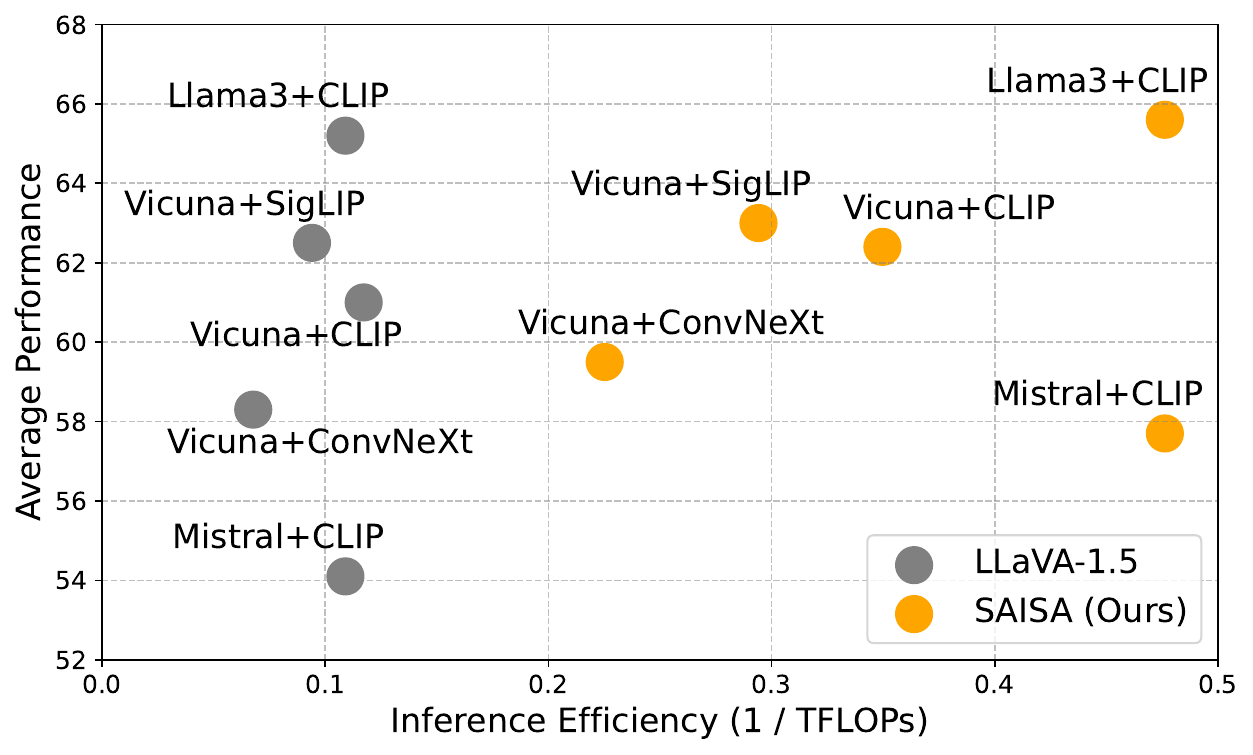} \\
\hfill
\includegraphics[width=0.935\linewidth]{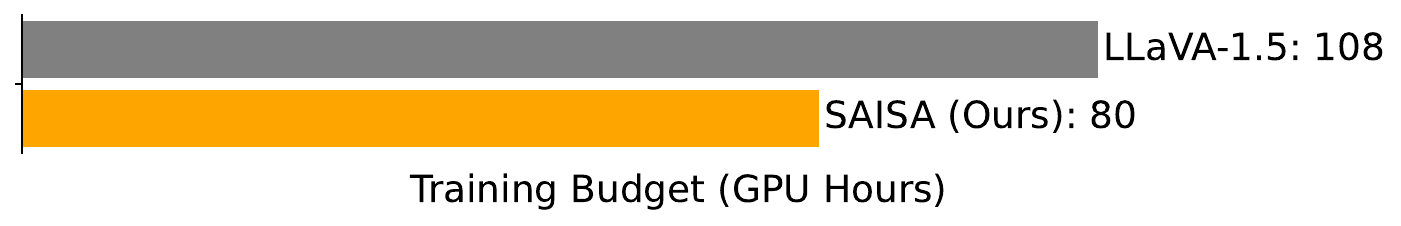}
\vspace{-0.1cm}
\caption{
\textbf{Top: Performance \textit{vs.} inference efficiency} based on various LLMs and visual encoders where Average Performance means an average of benchmark scores (MMMU, MMBench, MMBench-CN, POPE, GQA, SchienceQA IMG and OK-VQA) and inference efficiency is the inverse of inference TFLOPs.
When trained on the \textbf{same data} and using the \textbf{same number of visual tokens}, SAISA (\textcolor{orange}{orange}) offers a more favorable balance between inference efficiency and performance than LLaVA-1.5 (\textcolor{gray}{gray}).
\textbf{Bottom: Training budget comparison between SAISA and LLaVA-1.5} where we report the training GPU hours, using Vicuna-7B as LLM and CLIP-ViT-L/14-336 as visual encoder.
SAISA achieves higher training efficiency.
}
\label{fig:fig1}
\end{figure}

Multimodal Large Language Models (MLLMs)~\cite{gpt4v, liu2024improvedbaselinesvisualinstruction, Qwen-VL, chen2023minigptv2largelanguagemodel, dai2023instructblipgeneralpurposevisionlanguagemodels} have shown impressive capabilities in understanding and processing visual information.
They typically build on pre-trained Large Language Models (LLMs)~\cite{openai2024gpt4technicalreport, vicuna, bai2023qwentechnicalreport,  touvron2023llama2openfoundation} and align visual features with the LLMs.
There are two primary architectures for aligning visual and text modalities: embedding space alignment and cross-attention space alignment.
Embedding space alignment, e.g., LLaVA~\cite{liu2023visualinstructiontuning, liu2024improvedbaselinesvisualinstruction}, introduces a projector to align visual features with the LLM embedding space and feeds the visual and text tokens into the LLM.
Cross-attention space alignment, e.g., Flamingo~\cite{alayrac2022flamingovisuallanguagemodel}, inserts cross-attention blocks and aligns visual features with the attention spaces of these blocks.

However, despite the promising performance of these MLLMs, they involve a trade-off between training and inference efficiency.
On the one hand, MLLMs with embedding space alignment exhibit training efficiency, since they introduce only a small number of new parameters to the pre-trained LLMs.
For example, LLaVA-1.5-7B is trained in 108 GPU hours from Vicuna-7B~\cite{vicuna}.
However, this architecture significantly increases the number of input tokens, and the computational cost of self-attention grows quadratically with the number of tokens, leading to inefficiency during inference.
On the other hand, MLLMs with cross-attention space alignment achieve inference efficiency, since they do not require unrolling visual tokens, but they are inefficient during training for introducing a large number of new parameters to the pre-trained LLM.
In this paper, we take a step towards building MLLMs with efficiency during both training and inference.

In this paper, we perform a thorough analysis of these two architectures, identifing key factors for building MLLMs with both training and inference efficiency.
To optimize training efficiency, the key factor is minimizing the number of new parameters and employ modules in the pre-trained LLMs for interaction between visual and text modalities.
For improving inference efficiency, the main focus is reducing the computational costs associated with visual tokens, particularly in attention blocks and feed-forward networks (FFNs).
Building on the analysis of these factors, we introduce NAAViT (\textbf{N}o \textbf{A}ttention \textbf{A}mong \textbf{Vi}sual \textbf{T}okens), a self-attention mechanism which eliminates attention among visual tokens to enhance efficiency.
Since attention among visual tokens contributes significantly to the quadratically growing computational cost in self-attention blocks, we investigate whether this type of attention is truly essential for MLLMs.
Our pilot experiment on LLaVA-1.5 demonstrates that NAAViT outperforms vanilla self-attention, indicating that attention among visual tokens is highly redundancy.

Based on the findings above, we introduce SAISA (\textbf{S}elf-\textbf{A}ttention \textbf{I}nput \textbf{S}pace \textbf{A}lignment), an architecture for MLLMs with efficiency during both training and inference.
As illustrated in Figure~\ref{fig:saisa}(c), SAISA employs NAAViT self-attention blocks for multimodal interaction and directly aligns visual features with the input spaces of these blocks.
SAISA not only reduces the computational overhead of self-attention blocks but also significantly lowers the computational cost of FFNs by eliminating the need to apply FFNs to visual tokens.
We train SAISA on the same data as LLaVA-1.5 and validate its effectiveness on various LLMs and visual encoders.
As shown in Figure~\ref{fig:fig1}, SAISA outperforms LLaVA-1.5 in terms of performance, training efficiency, and inference efficiency.
Using Vicuna-7B-v1.5~\cite{vicuna} as LLM and CLIP-ViT-L/14-336~\cite{radford2021learningtransferablevisualmodels} as visual encoder, SAISA reduces training budget by 26\% and inference FLOPs by 66\%, while delivering superior performance.
Moreover, SAISA is orthogonal to and compatible with projectors to down-sample visual tokens, e.g. perceiver resampler~\cite{jaegle2021perceivergeneralperceptioniterative} and Q-Former~\cite{li2023blip2bootstrappinglanguageimagepretraining, dai2023instructblipgeneralpurposevisionlanguagemodels}, and visual token pruning mechanisms, e.g. FastV~\cite{chen2024imageworth12tokens} and LLaVA-PruMerge~\cite{shang2024llavaprumergeadaptivetokenreduction}.

We summarize our contributions as follows.
\begin{itemize}
    \item Based on our analysis of current MLLM architectures, we propose NAAViT to enhance efficiency of MLLMs, revealing the redundancy of self-attention in MLLMs.
    \item We introduce SAISA, an architecture for building MLLMs with both training and inference efficiency by eliminating the computational cost of attention among visual tokens and FFNs on visual tokens. 
    \item We validate the effectiveness of SAISA through comprehensive ablation studies across various LLMs and visual encoders.
\end{itemize}

%% file: sec/2_current.tex
\section{Analyzing Current MLLM Architectures}
\label{sec:architectures}
In this section, we analyze the two most common architectures to align visual features with the language model, and summarize key factors for building efficient MLLMs.

\begin{figure*}
    \centering
    \includegraphics[width=0.95\linewidth]{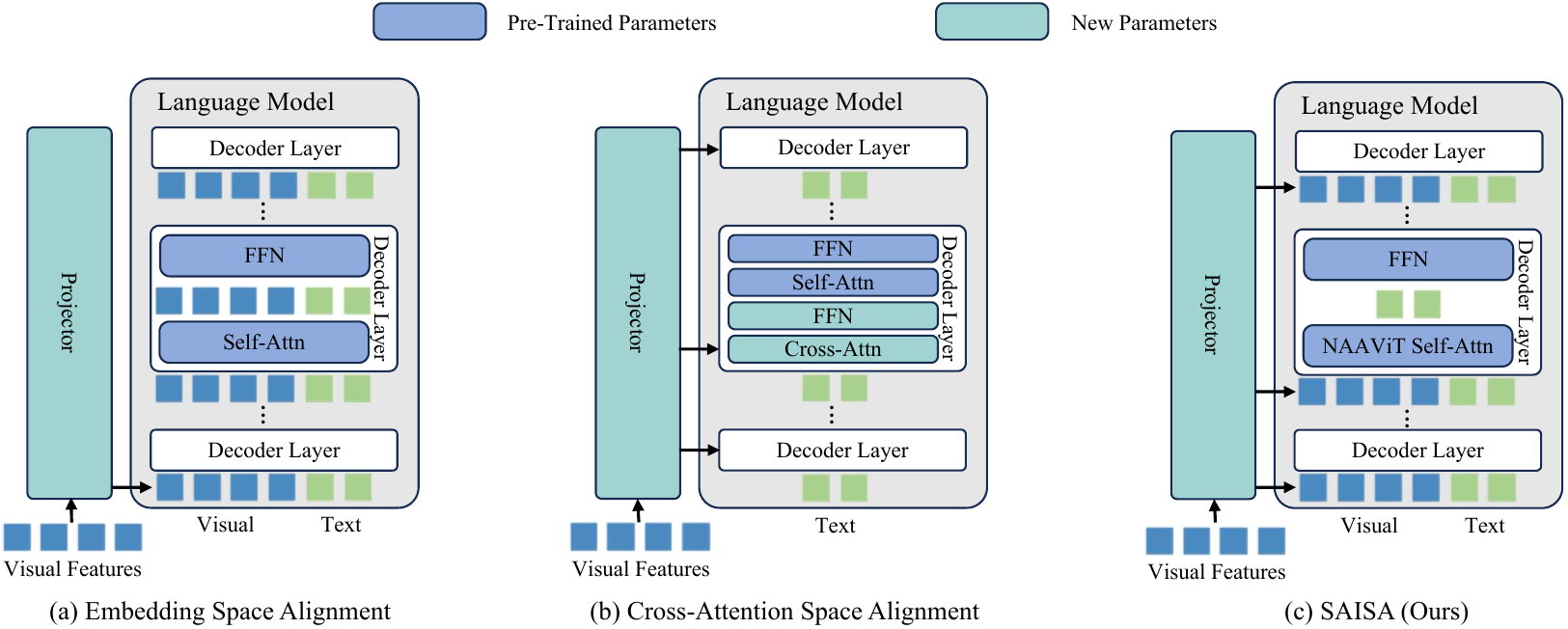}
    \vspace{-0.1cm}
    \caption{\textbf{Overview of SAISA and the mainstream architectures to align visual features with language model.} (a) Aligning visual features with the embedding space of the language model is inefficient during inference, e.g. LLaVA series. (b) Aligning visual features with the attention spaces of new cross-attention blocks is inefficient during training, e.g. Flamingo and OpenFlamingo. (c) SAISA aligns visual features with the self-attention input spaces of the language models, achieving efficiency during both training and inference.
    }
  \label{fig:saisa}
\end{figure*}

\vspace{-0.35cm}
\paragraph{Embedding Space Alignment.}
As illustrated in Figure~\ref{fig:saisa}(a), models with this architecture introduce a projector to align visual tokens with the text token embedding space.
They concatenate the aligned visual tokens and text tokens, and then feed them into the LLM.
Notable models with this architecture include  LLaVA-1.5~\cite{liu2024improvedbaselinesvisualinstruction}, Qwen-VL~\cite{Qwen-VL}, BLIP-2~\cite{li2023blip2bootstrappinglanguageimagepretraining}, InstructBLIP~\cite{dai2023instructblipgeneralpurposevisionlanguagemodels}, MiniGPT-4~\cite{zhu2023minigpt4enhancingvisionlanguageunderstanding} and MiniGPT-v2~\cite{chen2023minigptv2largelanguagemodel}.
These models introduce only a small number of new parameters to the pre-trained LLMs, allowing training MLLMs from the LLMs with minimal budget.

However, the concatenated token sequence leads to inefficiency during inference.
When the number of visual and text tokens is $v$ and $t$ respectively, the computational complexity of self-attention in LLM is $\mathcal{O}((v+t)^2)$.
This complexity consists of three components, $\mathcal{O}(v^2)$ for the attention among visual tokens, $\mathcal{O}(vt)$ for the interaction between text and visual tokens, and $\mathcal{O}(t^2)$ for the attention among text tokens.
Typically, MLLMs use hundreds or even thousands of visual tokens.
For examples, LLaVA-1.5 uses 576 visual tokens for a single image, and Sphinx~\cite{lin2023sphinxjointmixingweights} uses 2,890 visual tokens.
In contrast, the number of text tokens is much smaller in most VL tasks.
The average numbers of text tokens in MMMU~\cite{yue2024mmmumassivemultidisciplinemultimodal}, POPE~\cite{li2023evaluatingobjecthallucinationlarge} and ScienceQA IMG~\cite{lu2022learnexplainmultimodalreasoning} are 142, 68, and 210, respectively.
As a result, the attention among visual tokens dominates the quadratic computational overhead.
Moreover, since the FFNs of the LLM have a large hidden layer dimension, applying FFNs to visual tokens also brings substantial computational costs.

\vspace{-0.35cm}
\paragraph{Cross-Attention Space Alignment.}
As illustrated in Figure~\ref{fig:saisa}(b), models with this architecture insert cross-attention blocks and FFNs into the language model, and align visual features with the attention spaces of these cross-attention blocks.
Notable models with this architecture include Flamingo~\cite{alayrac2022flamingovisuallanguagemodel} and OpenFlamingo~\cite{awadalla2023openflamingoopensourceframeworktraining}.
In these models, the attention operation consists of only two components, the interaction between text and visual modalities with complexity $\mathcal{O}(vt)$ in the cross-attention blocks, and the attention among text tokens with complexity $\mathcal{O}(t^2)$ in the self-attention blocks.
Compared to embedding space alignment, there is no attention among visual features in the language model.
By not executing attention among visual tokens and not applying FFNs to visual tokens, these models are more efficient during inference.

However, the inserted cross-attention blocks and FFNs introduce a large number of new parameters to the pre-trained language model.
As a result, training an MLLM with this architecture requires a large amount of data.
For example, OpenFlamingo-9B adds 2 billion parameters to Llama-7B~\cite{touvron2023llamaopenefficientfoundation}, and requires training data with 180M samples.
In terms of model capabilities, previous work~\cite{dai2024nvlmopenfrontierclassmultimodal} finds that models utilizing this architecture perform worse than those using embedding space alignment when trained on the same data.

Based on the analyses above, we summarize the key factors for building efficient MLLMs as follows:
\begin{itemize}
    \item Reducing the number of new parameters and employing pre-trained modules for multimodal interaction lead to efficiency during training.
    \item Reducing computations related to visual tokens, including those in attention blocks and FFNs, leads to efficiency during inference.
\end{itemize}

%% file: sec/3_naavit.tex
\section{No Attention Among Visual Tokens}
\label{sec:rethinking}
In this section, we propose NAAViT (No Attention Among Visual Tokens) self-attention and perform a pilot experiment to investigate whether attention among visual tokens is necessary for MLLMs.
\subsection{Preliminary}
\paragraph{Vanilla Self-Attention.}
In a self-attention block, the input visual-text token sequence is formed as $X=[V,T]\in \mathbb{R}^{(v+t) \times h}$,
where $[\cdot,\cdot]$ denotes concatenation along the sequence dimension. To derive the query, key, and value representations, three linear layers are applied to obtain $X_q$,  $X_k$, and  $X_v$, respectively:
\begin{equation}
X_q=[V,T]W_Q, X_k=[V,T]W_K, X_v=[V,T]W_V
\end{equation}
Then, the attention operation is executed as:
\begin{equation}
\text{Attention}(X) = \text{softmax}\left(\frac{X_q X_k^T}{\sqrt{d}}\right) X_v\in \mathbb{R}^{(v+t) \times h}
\label{eq:attention}
\end{equation}
Typically, a causal attention mask is applied and queries can only attend to keys preceding them in the sequence.
The outputs are multiplied by another linear layer $W_O$ to update the hidden states through a residual connection:
\begin{equation}
\text{SA}(X) = \text{Attention}(X)W_O+X
\label{eq:2}
\end{equation}

\subsection{NAAViT}
As analyzed in Section~\ref{sec:architectures}, embedding space alignment achieves superior performance and training efficiency, making it the most popular architecture in MLLMs recently.
However, this architecture is inefficient during inference, largely attributable to the computational overhead of self-attention among visual tokens. In contrast, cross-attention space alignment does not perform attention among visual tokens in the language model, leading to inference efficiency.
A question arises here: \textbf{Is attention among visual tokens necessary for MLLMs?}

\begin{figure}[t]
  \centering
    \includegraphics[width=0.65\linewidth]{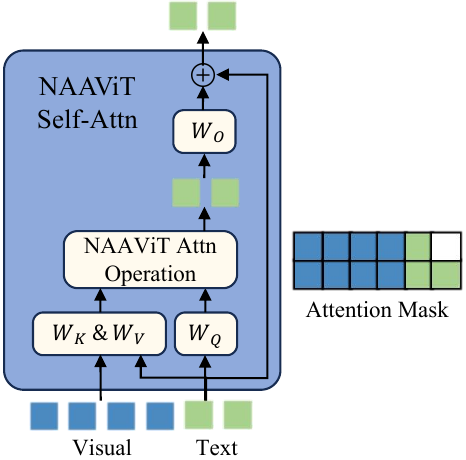}
    \vspace{-0.1cm}
   \caption{\textbf{NAAViT self-attention block.} NAAViT uses only text tokens as queries. The queries can attend to visual tokens and text tokens preceding them. Visual tokens are not updated in this block.}
   \label{fig:naavit}
\end{figure}

To answer this question, we propose NAAViT (No Attention Among Visual Tokens), which eliminates attention among visual tokens.
We illustrate the architecture of NAAViT in Figure~\ref{fig:naavit}.
Specifically, for the visual-text token sequence $X=[V,T]\in \mathbb{R}^{(v+t) \times h}$, queries $X_q$, keys $X_k$ and values $X_v$ are obtained as:
\begin{equation}
X_q=VW_Q, X_k=[V,T]W_K, X_v=[V,T]W_V
\end{equation}
The attention operation is executed as Equation~\ref{eq:attention}, but with NAAViT attention mask, where the queries can attend to visual tokens and text tokens preceding them.

When the number of visual and text tokens is $v$ and $t$ respectively, the vanilla self-attention exhibit a computational complexity of $\mathcal{O}((v+t)^2)$ for the attention operation in Equation~\ref{eq:attention}.
By eliminating the attention among visual tokens, NAAViT reduces the complexity to $\mathcal{O}(t(v+t))$.
Furthermore, since NAAViT reduces the query length from $v+t$ to $t$, it also significantly reduces computational costs associated with linear layers $W_Q$ and $W_O$.

\input{table/attention}

\subsection{Pilot Experiment}
We train a model under the same configurations as LLaVA-1.5-7B, but replace the vanilla self-attention blocks with NAAViT self-attention blocks.

In Table~\ref{tab:attention}, we compare NAAViT and vanilla self-attention on multiple MLLM benchmarks, including MMMU~\cite{yue2024mmmumassivemultidisciplinemultimodal}, 
MMBench~\cite{liu2024mmbenchmultimodalmodelallaround}, MMBench-CN~\cite{liu2024mmbenchmultimodalmodelallaround}, POPE~\cite{li2023evaluatingobjecthallucinationlarge}, ScienceQA IMG~\cite{lu2022learnexplainmultimodalreasoning} and OK-VQA~\cite{marino2019okvqavisualquestionanswering}.
Despite eliminating the attention among visual tokens, the model employing NAAViT outperforms the one using vanilla self-attention. 
Given that NAAViT substantially reduces computational overhead, it offers a favorable balance between performance and efficiency.

In conclusion, attention among visual tokens is highly redundant for building MLLMs.
In the following section, we introduce SAISA (Self-Attention Input Space Alignment) for efficient MLLMs based on NAAViT.

%% file: table/attention.tex
\begin{table}[t]
  \centering
    \scalebox{0.8}
    {
    \begin{tabular}{l|cccccc}
    \toprule
    \multirow{2}[2]{*}{Attention}  & MMMU & \multicolumn{2}{c}{MMBench} & \multirow{2}[2]{*}{POPE}  & SQA & OK- \\
        & VAL & EN & CN        &             & IMG & VQA \\
    \midrule
    Vanilla  & 35.7  & 64.3 & 58.3  & 86.8  & 66.8 & 53.4  \\
    \rowcolor{cyan!20} NAAViT & 36.0 & 64.9 & 58.4 & 86.9 & 68.7 & 56.0  \\
    \bottomrule
    \end{tabular}
    }
    \caption{\textbf{Effects of multimodal attention mechanisms.}
    NAAViT outperforms vanilla self-attention.
    }
  \label{tab:attention}
\end{table}

%% file: sec/4_method.tex
\section{SAISA}
\label{sec:4}

\begin{figure*}
  \centering
  \begin{subfigure}{0.49\linewidth}
    \includegraphics[height=4.5cm, width=\linewidth]{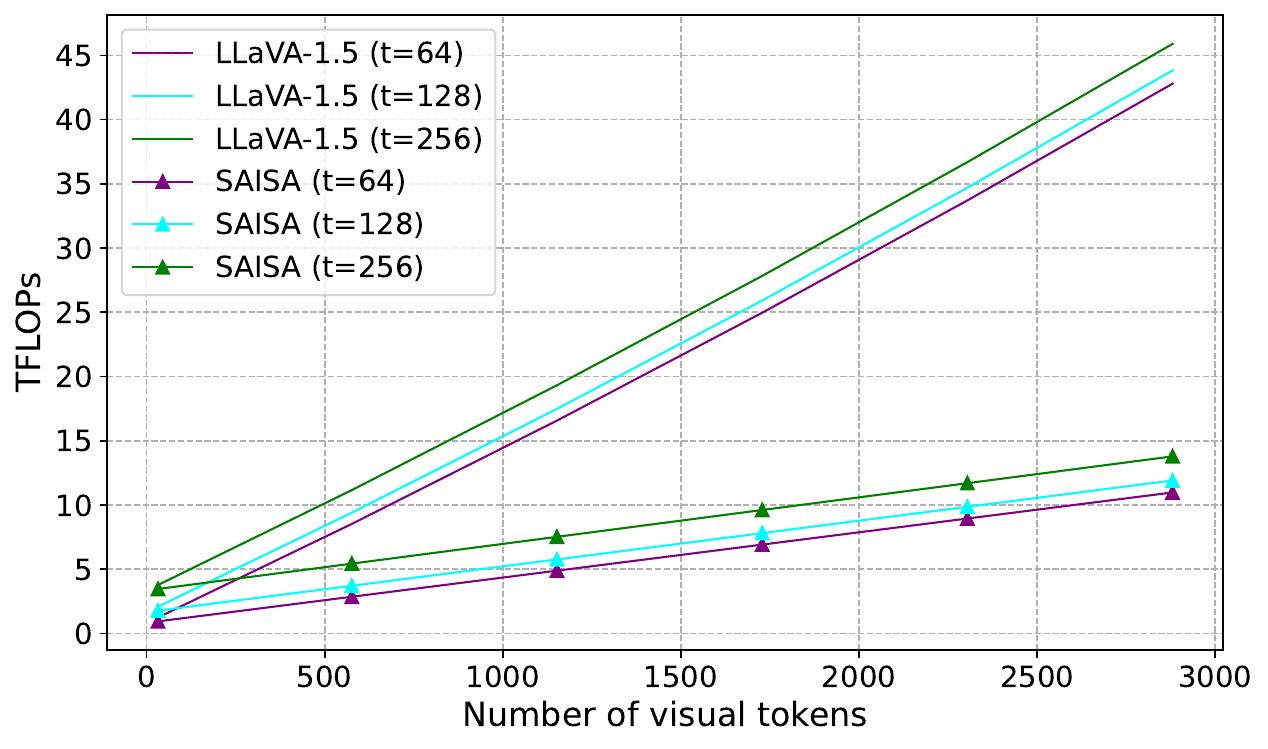}
    \caption{TFLOPs with different numbers of tokens}
    \label{fig:short-a}
  \end{subfigure}
  \hfill
  \begin{subfigure}{0.49\linewidth}
    \includegraphics[height=4.5cm, width=\linewidth]{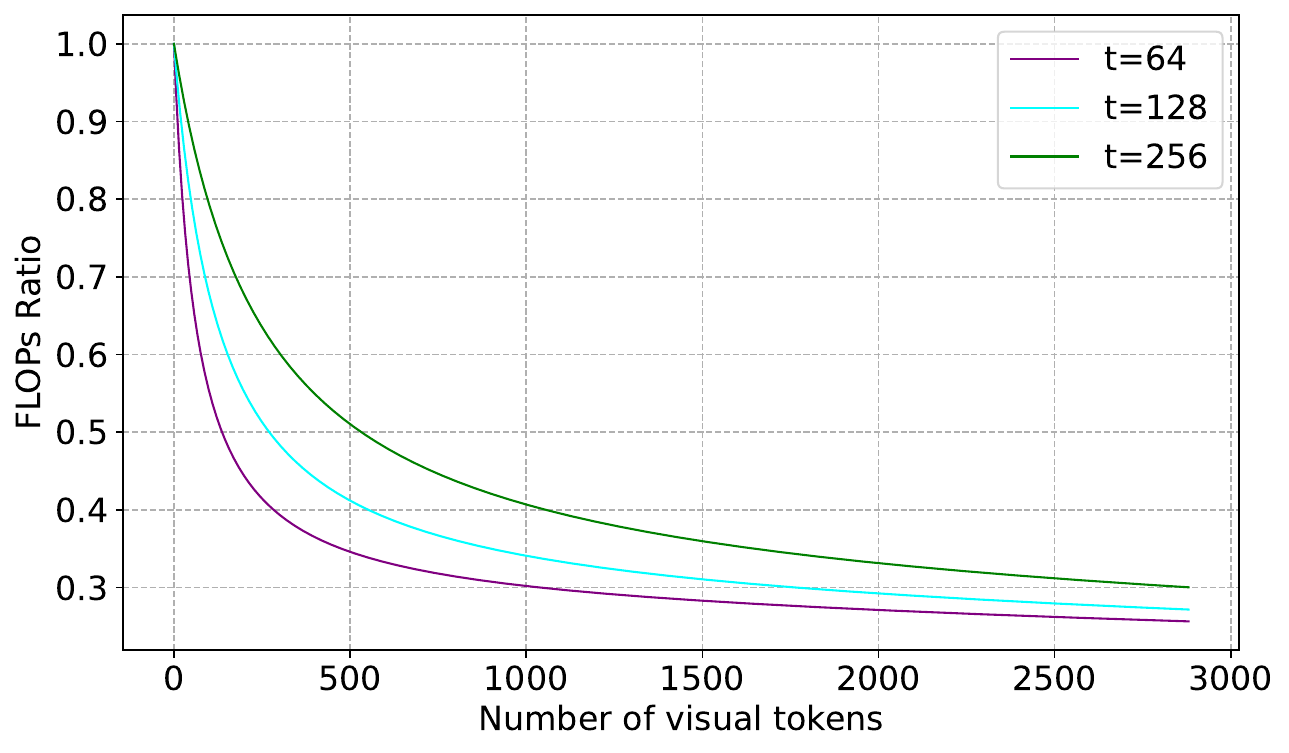}
    \caption{FLOPs ratio of SAISA and LLaVA-1.5 with different numbers of tokens}
    \label{fig:flops}
  \end{subfigure}
  
  \caption{\textbf{Inference computational costs comparison between SAISA and LLaVA-1.5} with different numbers of visual and text tokens, where t denotes the number of text tokens.
  SAISA achieves higher computational efficiency than LLAVA-1.5.}
  \label{fig:flops}
\end{figure*}

\subsection{Architecture}
As mentioned earlier, besides the attention operation, another factor that contributes to inference inefficiency is applying FFNs to visual tokens.
Based on NAAViT, which eliminates the attention among visual tokens, we propose SAISA (Self-Attention Input Space Alignment), an architecture for further enhancing MLLM efficiency.
In SAISA, we also eliminate FFNs' computations on visual tokens.

We illustrate the SAISA architecture in Figure~\ref{fig:saisa}(c).
SAISA contains a visual encoder to extract visual features, a projector, and an LLM.
Each layer of the LLM consists of a self-attention block and an FFN.
We utilize NAAViT in the self-attention blocks.
The purpose of the projector is to directly align the visual features with the input spaces of different self-attention blocks in the LLM.

Specifically, we assume $n$ is the number of layers in the LLM, $h$ is the hidden size of the LLM, $d$ is the dimension of visual features, $v$ is the number of visual tokens, and $t$ is the number of text tokens.

For an input image $I$, we first employ the visual encoder VE to extract visual features:
\begin{equation}
Z=\text{VE}(I)\in \mathbb{R}^{v \times d}
\end{equation}
Then, we use the projector P to directly align the visual features with the input spaces of different NAAViT self-attention blocks:
\begin{equation}
V=\text{P}(Z)\in \mathbb{R}^{n \times v \times h}
\end{equation}
For the $i$-th NAAViT self-attention block NAAViT$_i$, we input the aligned visual features $V_i\in \mathbb{R}^{v \times h}$ and hidden states of the text tokens $T_i\in \mathbb{R}^{t \times h}$:
\begin{equation}
H_i=\text{NAAViT}_i(V_i,T_i)\in \mathbb{R}^{t \times h}, i=1,2,3,\dots,n
\end{equation}
Notably, the NAAViT self-attention block only outputs the text tokens' hidden states $H_i$ for the subsequent FFN, and we apply the FFN to update the text tokens:
\begin{equation}
T_{i+1}=\text{FFN}_i(H_i)\in \mathbb{R}^{t \times h}, i=1,2,3,\dots,n
\end{equation}

\subsection{Projector}
Since each LLM layer operates in distinct self-attention spaces, the projector must flexibly align the visual features with each of the spaces.
For simplicity, we employ distinct two-layer MLPs for each layer of the LLM. 
When the LLM has $n$ layers, the projector contains $n$ MLPs.
Following LLaVA-1.5~\cite{liu2024improvedbaselinesvisualinstruction}, we set the intermediate size of each MLP to be the same as the hidden state size of the LLM.

Specifically, for the $i$-th layer in the LLM, the projector is executed as:
\begin{equation}
V_i=\text{MLP}_i(Z)= \varphi(Z W_{i,1}) W_{i,2}, i=1,2,3,\dots,n
\end{equation}
where $Z$ is the visual features from the visual encoder, $\varphi$ is the activation function like GELU~\cite{hendrycks2023gaussianerrorlinearunits}, $W_{i,1}\in \mathbb{R}^{d \times h}$ and $W_{i,2}\in \mathbb{R}^{h \times h}$ are the weight matrices of the two fully connected layers.

\subsection{Training Procedure}
The training procedure of SAISA consists of two stages: pre-training and fine-tuning.

\vspace{-0.35cm}
\paragraph{Pre-training.}
The objective of pre-training is to transform an LLM into an MLLM with a foundational comprehension of images, providing an initialization for the multimodal fine-tuning stage.
Following LLaVA-1.5, only the multimodal projector is trainable during this stage.
To improve training efficiency, we further reduce the number of trainable parameters.
Specifically, we train a shared MLP for all layers of the LLM.

\vspace{-0.35cm}
\paragraph{Fine-tuning.}
The objective of fine-tuning is to enable the model to follow visual instructions from users.
As an initialization, we replicate the pre-trained shared MLP $N$ times to initialize the MLPs in the projector, where $N$ denotes the number of layers of the LLM.
Following LLaVA-1.5, we utilize visual instruction data to train the model, and both the LLM and the projector are trainable during this stage.

\input{table/performance}
\input{table/vqa}

\subsection{Comparison of Computational Cost}
In this section, we compare the computational costs of SAISA and LLaVA-1.5.
We consider the computations of the LLM and the projector, as the computations of the visual encoder are identical in comparison.
We consider the computations of the self-attention blocks and the FFNs in the LLM.

For LLM with $n$ layers, we assume $h$ is the hidden state size, $m$ is the intermediate size of the FFNs, $t$ is the number of text tokens, and $v$ is the number of visual tokens.
To comprehensively consider LLMs with and without grouped query attention (GQA)~\cite{ainslie2023gqatraininggeneralizedmultiquery}, we assume $k$ is the output dimension of key/value matrices.
For the projector, we assume $d$ is the dimension of the input visual features.

For LLaVA-1.5, the FLOPs of the LLM are calculated as $2n(t+v)h(2h+3m+2k)+4n(t+v)^2h$, and the FLOPs of the projector are $2vhd+2vh^2$. The total FLOPs are $2n(t+v)h(2h+3m+2k)+4N(t+v)^2h+2vhd+2vh^2$

For SAISA, visual tokens are multiplied only by key and value matrices, and the key-value sequence length is $v$ in the attention operation.
Therefore, the FLOPs of the LLM can be estimated by $2nth(2h+3m+2k)+4nvhk+4nt(t+v)h$.
The FLOPs of the projector are $2nvhd+2nvh^2$.
The total FLOPs are $2nth(2h+3m+2k)+4nvhk+4nt(t+v)h+2nvhd+2nvh^2$.

Figure~\ref{fig:flops} compares the FLOPs of SAISA and LLaVA-1.5 with different numbers of tokens, based on Vicuna-7B-v1.5~\cite{vicuna}.
SAISA achieves a higher computational efficiency than LLAVA-1.5 when processing the same numbers of tokens.
For example, SAISA's FLOPs are only 34\% of LLaVA-1.5's when using CLIP-ViT-L/14-336~\cite{radford2021learningtransferablevisualmodels} as visual encoder and the number of text tokens is 64.

%% file: table/performance.tex
\begin{table*}[t]
  \centering
  {
  \renewcommand{\arraystretch}{1.0}
 
 \begin{tabular}{ll|c|c|c|c|c|cc|c}
    \toprule
    \multicolumn{1}{l}{\multirow{2}[2]{*}{Method}} & \multirow{2}[2]{*}{LLM} & \#Vis. & Training & Inference & MMMU & \multirow{2}[2]{*}{MME~\cite{fu2024mmecomprehensiveevaluationbenchmark}} & \multicolumn{2}{c|}{MMBench~\cite{liu2024mmbenchmultimodalmodelallaround}} & SEED~\cite{li2023seedbenchbenchmarkingmultimodalllms} \\
          &       & Tok. & Data$\downarrow$ & TFLOPs$\downarrow$ & VAL~\cite{yue2024mmmumassivemultidisciplinemultimodal} &       & EN & CN & Image \\
    \midrule
    BLIP-2 & Vicuna-7B & 32    & 129M  & 1.25  & -     & 1293.8 & -     & -     & - \\
    InstructBLIP & Vicuna-7B & 32    & 130M  & 1.25  & 30.6  & 1137.1 & 36.0  & 23.7  & 58.8 \\
    MiniGPT-4 & Vicuna-7B & 32    & 517k  & 1.25  & 23.6  & 770.6 & 32.7  & 11.9  & 31.4 \\
    MiniGPT-v2 & Llama2-7B & 256   & 326M   & 4.20  & 25.0  & 708.4 & 24.3  & -     & 29.4 \\
    Otter & Llama-7B & 64    & 2.1B  & 1.67  & -     & 1292.3 & 48.3  & -     & 35.2 \\
    Shikra & Vicuna-7B & 256   & 6.1M  & 4.20  & -     & -     & 58.8  & -     & - \\
    IDEFICS & Llama-7B & 64    & 354M  & 1.67  & 18.4  & 942.0  & 48.2  & 25.2  & 44.5 \\
    IDEFICS & Llama-65B & 64    & 354M  & 16.62 & -     & 1244.9     & 54.5  & 38.1  & 53.2 \\
    Qwen-VL-Chat & Qwen-7B & 256   & 1.4B  & 4.20  & 36.0  & 1435.2 & 60.6  & 56.7  & 65.4 \\
    LLaVA-1.5 & Vicuna-7B & 576   & 1.2M  & 8.53  & 35.7  & \textbf{1510.7} & 64.3  & 58.3  & \textbf{66.1} \\
    \rowcolor{cyan!20} SAISA (Ours) & Vicuna-7B & 576   & 1.2M  & 2.86  & \textbf{36.9} & 1461.9 & \textbf{65.7} & \textbf{59.0} & 64.5 \\
    \bottomrule
    \end{tabular}%
    }
  \caption{\textbf{Performance on comprehensive benchmarks for instruction-following MLLMs.}
  \#Vis. Tok.: the number of visual tokens involved in a single image.
  \#Training Data: accumulated multimodal pre-training and fine-tuning data volume.
  Inference TFLOPs: the computational cost of processing a single image when the number of text tokens is 64.
  $\downarrow$: a lower value in these columns is better.
    SAISA achieves the best performance on 3/5 benchmarks, while reducing inference TFLOPs by 66\% compared to LLaVA-1.5.
  }
  \label{tab:mllms}%
\end{table*}

%% file: table/vqa.tex
\begin{table*}[t]
  \centering
  {
  \renewcommand{\arraystretch}{1.0}
  \scalebox{0.95}
  { 
 \begin{tabular}{ll|c|cccc|cccc}
    \toprule
    \multirow{2}[2]{*}{Method} & \multirow{2}[2]{*}{LLM}  & Inference & \multicolumn{4}{c|}{POPE~\cite{li2023evaluatingobjecthallucinationlarge}}  & GQA & ScienceQA  & TextVQA & OK-VQA \\
          &        & TFLOPs$\downarrow$ & overall & rand & pop & adv      & \cite{hudson2019gqanewdatasetrealworld}      & IMG~\cite{lu2022learnexplainmultimodalreasoning}  & \cite{singh2019vqamodelsread} &  \cite{marino2019okvqavisualquestionanswering} \\
    \midrule
    \multicolumn{7}{l|}{\color{gray} PaLI-X-55B (Specialist SOTA, individually fine-tuned on each dataset) }                       & \textcolor{gray}{72.1} & \textcolor{gray}{-}  & \textcolor{gray}{71.4} & \textcolor{gray}{66.1}  \\
    \midrule
    Shikra & Vicuna-7B   & 4.20  & 84.7  & 86.9  & 84.0  & 83.1  & -     & -     & -     & - \\
    IDEFICS & Llama-7B   & 1.67  & 81.8  & 88.3  & 81.1  & 76.0    & 35.5  & 51.6  & 25.9 & 38.4 \\
    IDEFICS & Llama-65B   & 16.62 & 77.5  & 86.7  & 74.9  & 70.8    & 45.2  & 61.8  & 30.9 & 45.2 \\
    Qwen-VL-Chat & Qwen-7B   & 4.20  & 87.0  & \textbf{89.0} & 87.4  & 84.7    & 57.5  & 68.2  & \textbf{61.5} & 56.6 \\
    LLaVA-1.5 & Vicuna-7B   & 8.53  & 86.8  & 88.2  & 87.2  & \textbf{85.1}   & \textbf{62.0} & 66.8  & 58.2 & 53.4 \\
    \rowcolor{cyan!20} SAISA (Ours) & Vicuna-7B   & 2.86  & \textbf{87.2} & \textbf{89.0} & \textbf{87.6} & 85.0   & 60.9  & \textbf{70.1}  & 56.8 & \textbf{56.8} \\
    \bottomrule
    \end{tabular}
    }
  }
  \caption{
  \textbf{Performance on hallucination and visual question answering benchmarks.}
  We \textcolor{gray}{gray} out the specialist's method, which is individually fine-tuned on each dataset.
  }
  \label{tab:vqa}
\end{table*}

%% file: sec/5_experiments.tex
\section{Experiments}
In this section, we conduct comprehensive experiments to compare SAISA with existing MLLMs.
Furthermore, we perform a series of ablation experiments to further validate the effectiveness of the SAISA architecture.

\subsection{Setups}
\paragraph{Model Configuration.}
To make an apple-to-apple comparison between SAISA and LLaVA-1.5~\cite{liu2024improvedbaselinesvisualinstruction}, we use the same settings as LLaVA-1.5.
Specifically, we employ Vicuna-7B-v1.5~\cite{vicuna} as the default LLM and CLIP-ViT-L/14-336~\cite{radford2021learningtransferablevisualmodels} as the default visual encoder.

\vspace{-0.35cm}
\paragraph{Training Details.}
We utilize the same training data as LLaVA-1.5.
During pre-training, we adopt the pre-train dataset with 558k samples from LLaVA~\cite{liu2023visualinstructiontuning}. This stage takes around 1.5 hours on 8 A800 (80G) GPUs. 
During fine-tuning, we use the mixture instruction-following dataset from LLaVA-1.5. The dataset contains 665k samples from LLaVA-Instruct~\cite{liu2023visualinstructiontuning}, ShareGPT~\cite{sharegpt2023}, VQAv2, GQA~\cite{hudson2019gqanewdatasetrealworld}, OK-VQA~\cite{marino2019okvqavisualquestionanswering}, OCR-VQA~\cite{mishra2019ocrvqa}, A-OKVQA~\cite{schwenk2022aokvqabenchmarkvisualquestion}, TextCaps~\cite{sidorov2020textcapsdatasetimagecaptioning}, RefCOCO~\cite{kazemzadeh2014referitgame, mao2016generationcomprehensionunambiguousobject}, and Visual Genome~\cite{krishna2016visualgenomeconnectinglanguage}.
This stage takes around 8.5 hours on 8 A800 (80G) GPUs.
The total training budget  of SAISA is approximately 80 GPU hours.

\input{table/ablation}
\input{table/vision-centric}
\input{table/latency}


\subsection{Main Results}
The performance on benchmarks is shown in Table~\ref{tab:mllms}, Table~\ref{tab:vqa} and Table~\ref{tab:vision-centric}, and the result of the inference latency test is shown in Table~\ref{tab:latency}.

We evaluate SAISA on a range of benchmarks, including: (1) comprehensive benchmarks for instruction-following MLLMs such as MMMU~\cite{yue2024mmmumassivemultidisciplinemultimodal}, MME~\cite{fu2024mmecomprehensiveevaluationbenchmark}, MMBench~\cite{liu2024mmbenchmultimodalmodelallaround}, MMBench-CN~\cite{liu2024mmbenchmultimodalmodelallaround}, and SEED-bench~\cite{li2023seedbenchbenchmarkingmultimodalllms}; (2) hallucination benchmark such as POPE~\cite{li2023evaluatingobjecthallucinationlarge}, which evaluates MLLMs' degree of hallucination on three subsets: random, popular, and adversarial; (3) general visual question answering benchmarks such as GQA~\cite{hudson2019gqanewdatasetrealworld} and ScienceQA IMG~\cite{lu2022learnexplainmultimodalreasoning}; (4) fine-grained visual question answering benchmarks such as OK-VQA~\cite{marino2019okvqavisualquestionanswering} and TextVQA~\cite{singh2019vqamodelsread}, OK-VQA requires fine-grained image understanding and spatial understanding, and TextVQA is an OCR-related benchmark; (5) vision-centric MLLM benchmarks such as MMVP~\cite{tong2024eyeswideshutexploring} and CV-Bench~\cite{tong2024cambrian1fullyopenvisioncentric}.
In the inference latency test, the latency is reported as the time of LLM prefilling during inference with varying numbers of text tokens.
Table~\ref{tab:mllms} shows the comparison on the comprehensive benchmarks for instruct-following MLLMs.
SAISA outperforms BLIP-2~\cite{li2023blip2bootstrappinglanguageimagepretraining}, InstructBLIP~\cite{dai2023instructblipgeneralpurposevisionlanguagemodels}, MiniGPT-4~\cite{zhu2023minigpt4enhancingvisionlanguageunderstanding}, MiniGPT-v2\cite{chen2023minigptv2largelanguagemodel}, Otter~\cite{li2023ottermultimodalmodelincontext}, Shikra~\cite{chen2023shikraunleashingmultimodalllms}, and IDEFICS~\cite{idefics} utilizing LLama-7B and LLama-65B~\cite{touvron2023llamaopenefficientfoundation} across all these benchmarks.
Compared to Qwen-VL-Chat~\cite{Qwen-VL} trained on data with 1.4B samples, SAISA performs better on 4 out of 5 benchmarks.
Compared to LLaVA-1.5~\cite{liu2024improvedbaselinesvisualinstruction}, SAISA performs better on 3 out of 5 benchmarks.
Table~\ref{tab:vqa} shows the comparison on the hallucination and visual question answering benchmarks, and Table~\ref{tab:vision-centric} shows the comparison on vision-centric MLLM benchmarks.
Since most previous models do not evaluate performance on vision-centric MLLM benchmarks, we compare SAISA with LLaVA-1.5.
SAISA achieves the best overall performance compared to other baseline MLLMs, and strikes the optimal balance between effectiveness and efficiency.

\subsection{Ablation Study}
\paragraph{Ablation on LLMs and Visual Encoders.}
As presented in Table~\ref{tab:ablation}, we perform multiple ablation experiments on both LLMs and visual encoders to evaluate the robustness of SAISA.
We tune a set of SAISA models using a variety of LLM backbones and visual encoders.
The ablated LLMs include Vicuna-7B~\cite{vicuna} and two LLMs using grouped query attention (GQA)~\cite{ainslie2023gqatraininggeneralizedmultiquery}, such as Mistral-7B~\cite{jiang2023mistral7b} and Llama3-8B~\cite{llama3v}.
The ablated visual encoders include two ViT-based~\cite{dosovitskiy2021imageworth16x16words} visual backbones such as CLIP-ViT-L/14-336~\cite{radford2021learningtransferablevisualmodels} and  SigLIP-ViT-SO400M/14-384~\cite{zhai2023sigmoidlosslanguageimage}, and a ConvNeXt-based~\cite{liu2022convnet2020s} visual encoder such as ConvNeXt-XXL-1024 from OpenCLIP~\cite{ilharco_gabriel_2021_5143773, schuhmann2022laion5bopenlargescaledataset}.
The experimental results demonstrate that SAISA consistently achieves superior performance to LLaVA-1.5 across different LLM backbones and visual encoders, while dramatically reducing computational costs.

\vspace{-0.35cm}
\paragraph{Ablation on Pre-training Strategies.}
As shown in Table~\ref{tab:ablation_train}, we conduct an ablation study to investigate the effects of SAISA's pre-training strategies.
We tune a SAISA model where the full projector (32 MLPs) is tunable during pre-training, and the other settings keep the same as the original SAISA.
With more randomly initialized parameters, we observe a performance drop when pre-training the full projector.
We attribute this drop to the small amount of pre-training data with only 558k samples.
The ablation study demonstrates the effectiveness of our pre-training strategy, which provides a robust initialization for the subsequent fine-tuning stage.

\vspace{-0.35cm}
\paragraph{Ablation on Projector Designs.}
Previous works find that replacing linear projection with MLP projection improves performance in MLLM~\cite{liu2024improvedbaselinesvisualinstruction} and self-supervised learning~\cite{chen2020simpleframeworkcontrastivelearning, chen2020improvedbaselinesmomentumcontrastive}.
We conduct an experiment to investigate the impact of projector designs in SAISA.
We tune a model under the same configuration as the original SAISA model but replace the MLPs in the projector with linear layers.
Table ~\ref{tab:linear} shows that the model with MLPs in the projector performs better than the model with linear layers, which is consistent with the finding of the previous study~\cite{liu2024improvedbaselinesvisualinstruction}.
Notably, we note that even the SAISA model with linear layers achieves comparable performance to LLaVA-1.5 with MLP projection.
This observation provides additional evidence for the effectiveness of SAISA.

\input{table/ablattion_train}
\input{table/linear}

%% file: table/ablation.tex
\begin{table*}[t]
  \centering
  \resizebox{\textwidth}{!}
  {
  \renewcommand{\arraystretch}{1.0}
  { 
  \begin{tabular}{l|c|cc|c|ccccccc|c}
    \toprule
    \multirow{2}[2]{*}{Method} & \multirow{2}[2]{*}{LLM} & Visual & \#Vis. & Inference & MMMU & \multicolumn{2}{c}{MMBench} & \multirow{2}[2]{*}{POPE} & \multirow{2}[2]{*}{GQA} & SQA  & OK-  & \multirow{2}[2]{*}{Average} \\
          &       & Encoder & Tok. & TFLOPs$\downarrow$ & VAL & EN & CN    &         &    & IMG  & VQA &  \\
    \midrule
    LLaVA-1.5 & Vicuna & SigLIP & 729   & 10.63 & 36.6  & 66.2 & 58.9  & 86.5 & 62.5   & \textbf{70.5}   & \textbf{56.4} & 62.5 \\
    \rowcolor{cyan!20} SAISA & Vicuna & SigLIP & 729   & 3.40  & \textbf{37.4}  & \textbf{67.5} & \textbf{60.7}  & \textbf{87.0}  & \textbf{62.9}  & 70.0  & 55.8 & \textbf{63.0} \\
    \midrule
    LLaVA-1.5 & Vicuna & Conv & 1024  & 14.76 & 34.6  & 56.6 & 49.6  & \textbf{88.2}  & \textbf{61.1}  & 66.4  & 51.4 & 58.3 \\
    \rowcolor{cyan!20} SAISA & Vicuna & Conv & 1024  & 4.44  & \textbf{35.1}  & \textbf{61.1} & \textbf{54.9}  & 87.0 & 57.7   & \textbf{66.5}  & \textbf{54.4} & \textbf{59.5} \\
    \midrule
    LLaVA-1.5 & Vicuna & CLIP  & 576   & 8.53  & 35.7  & 64.3 & 58.3 & 86.8  & \textbf{62.0}  & 66.8  & 53.4 & 61.0 \\
    \rowcolor{cyan!20} SAISA & Vicuna & CLIP  & 576   & 2.86  & \textbf{36.9}  & \textbf{65.7} & \textbf{59.0} & \textbf{87.2} & 60.9   & \textbf{70.1}  & \textbf{56.8} & \textbf{62.4} \\
    \midrule
    LLaVA-1.5 & Mistral & CLIP  & 576   & 9.17  & 34.8  & 65.9 & 54.9  & \textbf{87.2}  & \textbf{62.0}   & \textbf{71.6}  & 2.5* & 54.1 \\
    \rowcolor{cyan!20} SAISA & Mistral & CLIP  & 576   & 2.10  & \textbf{35.9}  & \textbf{67.5} & \textbf{57.5}  & 86.9 & 61.2   & 71.2  & \textbf{23.9}* & \textbf{57.7} \\
    \midrule
    LLaVA-1.5 & Llama3 & CLIP  & 576   & 9.17  & 36.8  & 70.4 & 64.2 & \textbf{87.2}  & \textbf{63.5}  & 73.3  & \textbf{61.2} & 65.2 \\
    \rowcolor{cyan!20} SAISA & Llama3 & CLIP  & 576   & 2.10  & \textbf{38.3}  & \textbf{71.3} & \textbf{65.2} & 86.8  & 61.8  & \textbf{74.4}  & 60.7 & \textbf{65.6} \\
    \bottomrule
    \end{tabular}
    }
  }
  \caption{
  \textbf{Ablation on LLMs and Visual Encoders.}
  Here, ``Vicuna" = Vicuna-7B, ``Mistral" = Mistral-7B, ``Llama3" = Llama3-8B, ``SigLIP" = SigLIP-ViT-SO400M/14-384, ``Conv" = OpenCLIP-ConvNeXt-XXL-1024, and ``CLIP" = CLIP-ViT-L/14-336.
  \#Vis. Tok. denotes the number of visual tokens involved in a single image.
  SAISA consistently demonstrates superior performance to LLaVA-1.5 across these LLMs and visual encoders, while dramatically reducing computational costs.
  *Both LLaVA-1.5 and SAISA models on Mistral 7B exhibit low performance on OK-VQA, because they respond to most questions in this benchmark with "Unanswerable".
  }
  \label{tab:ablation}
\end{table*}

%% file: table/vision-centric.tex
\begin{table}[t]
  \centering
    \scalebox{0.9}
    {
    \begin{tabular}{l|c|ccc}
    \toprule
    \multirow{2}[2]{*}{Method} & Inference &  MMVP  & \multicolumn{2}{c}{CV-Bench~\cite{tong2024cambrian1fullyopenvisioncentric}}     \\
         & TFLOPs$\downarrow$      &   \cite{tong2024eyeswideshutexploring}          & 2D & 3D  \\
    \midrule
    LLaVA-1.5 & 8.53 & 24.7    & \textbf{56.6}  & 59.5   \\
    \rowcolor{cyan!20} SAISA (Ours) & 2.86 & \textbf{26.0}  & 56.2  & \textbf{59.8}   \\
    \bottomrule
    \end{tabular}
    }
    \caption{\textbf{Performance on vision-centric MLLM benchmarks,} based on Vicuna and CLIP.}
  \label{tab:vision-centric}
\end{table}

%% file: table/latency.tex
\begin{table}[t]
  \centering
    \scalebox{0.8}
    {
    \begin{tabular}{l|ccc}
    \toprule
    \multirow{2}[2]{*}{Method}  &  64 Text Tok.  & 128 Text Tok.  & 256 Text Tok.   \\
         &  Latency/ms$\downarrow$     &  Latency/ms$\downarrow$           & Latency/ms$\downarrow$  \\
    \midrule
    LLaVA-1.5  & 56.0    & 64.4  & 75.6   \\
    \rowcolor{cyan!20} SAISA (Ours)  & 33.0  & 37.1  & 45.9   \\
    \bottomrule
    \end{tabular}
    }
    \caption{\textbf{Result of the inference latency test,} based on Vicuna and CLIP, where the number of visual tokens is 576. 
    }
  \label{tab:latency}
\end{table}

%% file: table/ablattion_train.tex
\begin{table}[t]
  \centering
    \resizebox{\linewidth}{!}
    {
    \begin{tabular}{l|cccccc}
    \toprule
    Pre-trained &  MMMU & \multicolumn{2}{c}{MMBench} & \multirow{2}[2]{*}{POPE}  & SQA & OK-  \\
      Parameters  & VAL      &  EN & CN &           & IMG & VQA \\
    \midrule
    Full Projector & 34.8 & 59.2 & 51.1   & 85.6  & 67.8 & 53.1 \\
    \rowcolor{cyan!20} Shared MLP & 36.9 & 65.7 & 59.0 & 87.2  & 70.1 & 56.8  \\
    \bottomrule
    \end{tabular}
    }
    \caption{\textbf{Ablation on Pre-training Strategies.} 
    ``Full Projector" denotes we pre-train 32 MLPs.
    ``Shared MLP" denotes our strategy, which involves tuning a shared MLP for efficiency.
    Our strategy provides an effective initialization for visul fine-tuning when using the small pre-training dataset with 558k samples.
    }
  \label{tab:ablation_train}
\end{table}

%% file: table/linear.tex
\begin{table}[t]
  \centering
    \resizebox{\linewidth}{!}
    {
    \begin{tabular}{l|c|cccccc}
    \toprule
    \multirow{2}[2]{*}{Method} & \multirow{2}[2]{*}{Proj.} & MMMU & \multicolumn{2}{c}{MMBench} & \multirow{2}[2]{*}{POPE}  & SQA & OK-  \\
        &  & VAL  & EN & CN     &             & IMG & VQA \\
    \midrule
    LLaVA-1.5 & MLP & 35.7  & 64.3 & 58.3  & 86.8  & 66.8 & 53.4  \\
    SAISA & Linear & 35.7 & 65.3 & 56.6   & 85.8  & 69.2 & 53.6  \\
    \rowcolor{cyan!20} SAISA & MLP & 36.9 & 65.7 & 59.0 & 87.2  & 70.1 & 56.8  \\
    \bottomrule
    \end{tabular}
    }
    \caption{\textbf{Ablation on Projector Designs.} ``Proj." denotes projector type.
    The SAISA model that uses MLPs outperforms the model that uses linear layers.
    Notably, SAISA with linear layers achieves comparable performance to LLaVA-1.5 with MLP.
    }
  \label{tab:linear}
\end{table}

%% file: sec/6_related_work.tex
\section{Related Work}

\subsection{Multimodal Large Language Models}
Multimodal Large Language Models (MLLMs) are typically built on Large Language Models (LLMs)~\cite{vicuna, llama3v, jiang2023mistral7b, touvron2023llamaopenefficientfoundation, touvron2023llama2openfoundation, bai2023qwentechnicalreport} by aligning visual features generated by visual encoders~\cite{radford2021learningtransferablevisualmodels, zhai2023sigmoidlosslanguageimage, ilharco_gabriel_2021_5143773, liu2022convnet2020s} with the LLMs.
There are two most common architectures for this purpose, embedding space alignment and cross-attention space alignment.
For embedding space alignment~\cite{liu2023visualinstructiontuning, liu2024improvedbaselinesvisualinstruction, li2023blip2bootstrappinglanguageimagepretraining, dai2023instructblipgeneralpurposevisionlanguagemodels, Qwen-VL, chen2023minigptv2largelanguagemodel, zhu2023minigpt4enhancingvisionlanguageunderstanding}, MLLMs align visual features with the text token embedding space via a projector and concatenate the visual and text tokens as the LLM input.
These models exhibit efficiency during training due to the small number of new parameters but suffer from inference inefficiency because of the long token sequence.
For cross-attention space alignment~\cite{alayrac2022flamingovisuallanguagemodel, OpenFlamingov2, awadalla2023openflamingoopensourceframeworktraining}, MLLMs introduce new cross-attention blocks for the interaction between text and visual modalities, and align the visual features with the cross-attention spaces.
These models achieve efficiency during inference by eliminating the need to unroll visual tokens, but require a substantial amount of data to train the new parameters.
In this paper, we propose SAISA (Self-Attention Input Space Alignment), an architecture for building MLLMs with efficiency during both training and inference.

\subsection{Efficiency Optimization for MLLMs}
To reduce the computational cost of MLLMs, previous works mainly fall into two categories: model architecture and inference stage token reduction.
For model architecture, Qwen-VL-Chat~\cite{Qwen-VL}, LLaMA-VID~\cite{li2023llamavidimageworth2}, BLIP series~\cite{li2023blip2bootstrappinglanguageimagepretraining, dai2023instructblipgeneralpurposevisionlanguagemodels} and MiniGPT-4~\cite{zhu2023minigpt4enhancingvisionlanguageunderstanding} utilize attention-based mechanisms to down-sample visual tokens before they are fed into LLMs.
MemVP~\cite{jie2024memoryspacevisualpromptingefficient} and VLoRA~\cite{ma2024visualperceptionlargelanguage} integrate visual tokens into LLMs' weights through parameter-efficient fine-tuning (PEFT) such as LoRA~\cite{hu2021loralowrankadaptationlarge}, but overlook fine-grained visual information and retain limited spatial understanding.
EE-MLLM~\cite{ma2024eemllmdataefficientcomputeefficientmultimodal} introduces aligners to update visual tokens in the MLLM, but the aligners still involve substantial computational overhead.
For inference stage token reduction, FastV~\cite{chen2024imageworth12tokens} and LLaVA-PruMerge~\cite{shang2024llavaprumergeadaptivetokenreduction} focus on certain "anchor" tokens and prune the other visual tokens.
VoCo-LLaMA~\cite{ye2024vocollamavisioncompressionlarge} compresses vision tokens into compressed tokens.
In this paper, we investigate redundancy within the LLM architecture of the current mainstream MLLM~\cite{liu2024improvedbaselinesvisualinstruction} and propose SAISA, which eliminates redundant computations.
SAISA achieves efficiency and maintains fine-grained visual understanding by preserving the original number of visual tokens.
Notably, SAISA is orthogonal to and compatible with most previous methods~\cite{jaegle2021perceivergeneralperceptioniterative, li2023llamavidimageworth2, li2023blip2bootstrappinglanguageimagepretraining, chen2024imageworth12tokens, shang2024llavaprumergeadaptivetokenreduction, ye2024vocollamavisioncompressionlarge, Qwen-VL, dai2023instructblipgeneralpurposevisionlanguagemodels, zhu2023minigpt4enhancingvisionlanguageunderstanding}.

%% file: sec/7_conclusion.tex
\section{Conclusion}
In this paper, we take a step towards developing multimodal large language models (MLLMs) with efficiency during both training and inference.
To achieve this, we conduct a study of current MLLM architectures and find the key factors for building efficient MLLMs.
By integrating these factors and gradually reducing redundant computations, we propose \textbf{SAISA}, an effective and efficient architecture for MLLMs.
SAISA demonstrates the ability to dramatically reduce the computational costs of existing MLLMs without compromising their capabilities.

\vspace{-0.35cm}
\paragraph{Limitations.}
Despite the promising performance and efficiency of SAISA, several limitations must be acknowledged.
First, SAISA employs a projector with distinct MLPs for each layer of the LLM backbone for simplicity, introducing a number of parameters that cannot be ignored.
The development of a more efficient and effective projector could facilitate the advancement of more powerful MLLMs.
Second, SAISA is not yet capable of processing more complicated visual information, such as high-resolution images, multiple images, interleaved text-and-image content, and videos.
Third, SAISA is not yet capable of processing more modalities such as audio.
Finally, SAISA's capabilities are limited in following complex real-world instructions and solving problems in specific domains such as medicine.
These capabilities could be improved by scaling up pre-training and fine-tuning with high-quality, domain-specific data.

%% file: main.bbl
\begin{thebibliography}{60}
\providecommand{\natexlab}[1]{#1}
\providecommand{\url}[1]{\texttt{#1}}
\expandafter\ifx\csname urlstyle\endcsname\relax
  \providecommand{\doi}[1]{doi: #1}\else
  \providecommand{\doi}{doi: \begingroup \urlstyle{rm}\Url}\fi

\bibitem[Ainslie et~al.(2023)Ainslie, Lee-Thorp, de~Jong, Zemlyanskiy, Lebrón, and Sanghai]{ainslie2023gqatraininggeneralizedmultiquery}
Joshua Ainslie, James Lee-Thorp, Michiel de Jong, Yury Zemlyanskiy, Federico Lebrón, and Sumit Sanghai.
\newblock Gqa: Training generalized multi-query transformer models from multi-head checkpoints, 2023.

\bibitem[Alayrac et~al.(2022)Alayrac, Donahue, Luc, Miech, Barr, Hasson, Lenc, Mensch, Millican, Reynolds, Ring, Rutherford, Cabi, Han, Gong, Samangooei, Monteiro, Menick, Borgeaud, Brock, Nematzadeh, Sharifzadeh, Binkowski, Barreira, Vinyals, Zisserman, and Simonyan]{alayrac2022flamingovisuallanguagemodel}
Jean-Baptiste Alayrac, Jeff Donahue, Pauline Luc, Antoine Miech, Iain Barr, Yana Hasson, Karel Lenc, Arthur Mensch, Katie Millican, Malcolm Reynolds, Roman Ring, Eliza Rutherford, Serkan Cabi, Tengda Han, Zhitao Gong, Sina Samangooei, Marianne Monteiro, Jacob Menick, Sebastian Borgeaud, Andrew Brock, Aida Nematzadeh, Sahand Sharifzadeh, Mikolaj Binkowski, Ricardo Barreira, Oriol Vinyals, Andrew Zisserman, and Karen Simonyan.
\newblock Flamingo: a visual language model for few-shot learning, 2022.

\bibitem[Awadalla et~al.(2023)Awadalla, Gao, Gardner, Hessel, Hanafy, Zhu, Marathe, Bitton, Gadre, Sagawa, Jitsev, Kornblith, Koh, Ilharco, Wortsman, and Schmidt]{awadalla2023openflamingoopensourceframeworktraining}
Anas Awadalla, Irena Gao, Josh Gardner, Jack Hessel, Yusuf Hanafy, Wanrong Zhu, Kalyani Marathe, Yonatan Bitton, Samir Gadre, Shiori Sagawa, Jenia Jitsev, Simon Kornblith, Pang~Wei Koh, Gabriel Ilharco, Mitchell Wortsman, and Ludwig Schmidt.
\newblock Openflamingo: An open-source framework for training large autoregressive vision-language models, 2023.

\bibitem[Bai et~al.(2023{\natexlab{a}})Bai, Bai, Chu, Cui, Dang, Deng, Fan, Ge, Han, Huang, Hui, Ji, Li, Lin, Lin, Liu, Liu, Lu, Lu, Ma, Men, Ren, Ren, Tan, Tan, Tu, Wang, Wang, Wang, Wu, Xu, Xu, Yang, Yang, Yang, Yang, Yao, Yu, Yuan, Yuan, Zhang, Zhang, Zhang, Zhang, Zhou, Zhou, Zhou, and Zhu]{bai2023qwentechnicalreport}
Jinze Bai, Shuai Bai, Yunfei Chu, Zeyu Cui, Kai Dang, Xiaodong Deng, Yang Fan, Wenbin Ge, Yu Han, Fei Huang, Binyuan Hui, Luo Ji, Mei Li, Junyang Lin, Runji Lin, Dayiheng Liu, Gao Liu, Chengqiang Lu, Keming Lu, Jianxin Ma, Rui Men, Xingzhang Ren, Xuancheng Ren, Chuanqi Tan, Sinan Tan, Jianhong Tu, Peng Wang, Shijie Wang, Wei Wang, Shengguang Wu, Benfeng Xu, Jin Xu, An Yang, Hao Yang, Jian Yang, Shusheng Yang, Yang Yao, Bowen Yu, Hongyi Yuan, Zheng Yuan, Jianwei Zhang, Xingxuan Zhang, Yichang Zhang, Zhenru Zhang, Chang Zhou, Jingren Zhou, Xiaohuan Zhou, and Tianhang Zhu.
\newblock Qwen technical report, 2023{\natexlab{a}}.

\bibitem[Bai et~al.(2023{\natexlab{b}})Bai, Bai, Yang, Wang, Tan, Wang, Lin, Zhou, and Zhou]{Qwen-VL}
Jinze Bai, Shuai Bai, Shusheng Yang, Shijie Wang, Sinan Tan, Peng Wang, Junyang Lin, Chang Zhou, and Jingren Zhou.
\newblock Qwen-vl: A versatile vision-language model for understanding, localization, text reading, and beyond.
\newblock \emph{arXiv preprint arXiv:2308.12966}, 2023{\natexlab{b}}.

\bibitem[Chen et~al.(2023{\natexlab{a}})Chen, Zhu, Shen, Li, Liu, Zhang, Krishnamoorthi, Chandra, Xiong, and Elhoseiny]{chen2023minigptv2largelanguagemodel}
Jun Chen, Deyao Zhu, Xiaoqian Shen, Xiang Li, Zechun Liu, Pengchuan Zhang, Raghuraman Krishnamoorthi, Vikas Chandra, Yunyang Xiong, and Mohamed Elhoseiny.
\newblock Minigpt-v2: large language model as a unified interface for vision-language multi-task learning, 2023{\natexlab{a}}.

\bibitem[Chen et~al.(2023{\natexlab{b}})Chen, Zhang, Zeng, Zhang, Zhu, and Zhao]{chen2023shikraunleashingmultimodalllms}
Keqin Chen, Zhao Zhang, Weili Zeng, Richong Zhang, Feng Zhu, and Rui Zhao.
\newblock Shikra: Unleashing multimodal llm's referential dialogue magic, 2023{\natexlab{b}}.

\bibitem[Chen et~al.(2024)Chen, Zhao, Liu, Bai, Lin, Zhou, and Chang]{chen2024imageworth12tokens}
Liang Chen, Haozhe Zhao, Tianyu Liu, Shuai Bai, Junyang Lin, Chang Zhou, and Baobao Chang.
\newblock An image is worth 1/2 tokens after layer 2: Plug-and-play inference acceleration for large vision-language models, 2024.

\bibitem[Chen et~al.(2020{\natexlab{a}})Chen, Kornblith, Norouzi, and Hinton]{chen2020simpleframeworkcontrastivelearning}
Ting Chen, Simon Kornblith, Mohammad Norouzi, and Geoffrey Hinton.
\newblock A simple framework for contrastive learning of visual representations, 2020{\natexlab{a}}.

\bibitem[Chen et~al.(2020{\natexlab{b}})Chen, Fan, Girshick, and He]{chen2020improvedbaselinesmomentumcontrastive}
Xinlei Chen, Haoqi Fan, Ross Girshick, and Kaiming He.
\newblock Improved baselines with momentum contrastive learning, 2020{\natexlab{b}}.

\bibitem[Chiang et~al.(2023)Chiang, Li, Lin, Sheng, Wu, Zhang, Zheng, Zhuang, Zhuang, Gonzalez, Stoica, and Xing]{vicuna}
Wei-Lin Chiang, Zhuohan Li, Zi Lin, Ying Sheng, Zhanghao Wu, Hao Zhang, Lianmin Zheng, Siyuan Zhuang, Yonghao Zhuang, Joseph~E. Gonzalez, Ion Stoica, and Eric~P. Xing.
\newblock Vicuna: An open-source chatbot impressing gpt-4 with 90\%* chatgpt quality, 2023.

\bibitem[Dai et~al.(2023)Dai, Li, Li, Tiong, Zhao, Wang, Li, Fung, and Hoi]{dai2023instructblipgeneralpurposevisionlanguagemodels}
Wenliang Dai, Junnan Li, Dongxu Li, Anthony Meng~Huat Tiong, Junqi Zhao, Weisheng Wang, Boyang Li, Pascale Fung, and Steven Hoi.
\newblock Instructblip: Towards general-purpose vision-language models with instruction tuning, 2023.

\bibitem[Dai et~al.(2024)Dai, Lee, Wang, Yang, Liu, Barker, Rintamaki, Shoeybi, Catanzaro, and Ping]{dai2024nvlmopenfrontierclassmultimodal}
Wenliang Dai, Nayeon Lee, Boxin Wang, Zhuolin Yang, Zihan Liu, Jon Barker, Tuomas Rintamaki, Mohammad Shoeybi, Bryan Catanzaro, and Wei Ping.
\newblock Nvlm: Open frontier-class multimodal llms, 2024.

\bibitem[Dosovitskiy et~al.(2021)Dosovitskiy, Beyer, Kolesnikov, Weissenborn, Zhai, Unterthiner, Dehghani, Minderer, Heigold, Gelly, Uszkoreit, and Houlsby]{dosovitskiy2021imageworth16x16words}
Alexey Dosovitskiy, Lucas Beyer, Alexander Kolesnikov, Dirk Weissenborn, Xiaohua Zhai, Thomas Unterthiner, Mostafa Dehghani, Matthias Minderer, Georg Heigold, Sylvain Gelly, Jakob Uszkoreit, and Neil Houlsby.
\newblock An image is worth 16x16 words: Transformers for image recognition at scale, 2021.

\bibitem[Fu et~al.(2024)Fu, Chen, Shen, Qin, Zhang, Lin, Yang, Zheng, Li, Sun, Wu, and Ji]{fu2024mmecomprehensiveevaluationbenchmark}
Chaoyou Fu, Peixian Chen, Yunhang Shen, Yulei Qin, Mengdan Zhang, Xu Lin, Jinrui Yang, Xiawu Zheng, Ke Li, Xing Sun, Yunsheng Wu, and Rongrong Ji.
\newblock Mme: A comprehensive evaluation benchmark for multimodal large language models, 2024.

\bibitem[Hendrycks and Gimpel(2023)]{hendrycks2023gaussianerrorlinearunits}
Dan Hendrycks and Kevin Gimpel.
\newblock Gaussian error linear units (gelus), 2023.

\bibitem[Hu et~al.(2021)Hu, Shen, Wallis, Allen-Zhu, Li, Wang, Wang, and Chen]{hu2021loralowrankadaptationlarge}
Edward~J. Hu, Yelong Shen, Phillip Wallis, Zeyuan Allen-Zhu, Yuanzhi Li, Shean Wang, Lu Wang, and Weizhu Chen.
\newblock Lora: Low-rank adaptation of large language models, 2021.

\bibitem[Hudson and Manning(2019)]{hudson2019gqanewdatasetrealworld}
Drew~A. Hudson and Christopher~D. Manning.
\newblock Gqa: A new dataset for real-world visual reasoning and compositional question answering, 2019.

\bibitem[IDEFICS(2023)]{idefics}
IDEFICS.
\newblock Introducing idefics: An open reproduction of state-of-the-art visual language model.
\newblock \url{https://huggingface.co/blog/idefics}, 2023.

\bibitem[Ilharco et~al.(2021)Ilharco, Wortsman, Wightman, Gordon, Carlini, Taori, Dave, Shankar, Namkoong, Miller, Hajishirzi, Farhadi, and Schmidt]{ilharco_gabriel_2021_5143773}
Gabriel Ilharco, Mitchell Wortsman, Ross Wightman, Cade Gordon, Nicholas Carlini, Rohan Taori, Achal Dave, Vaishaal Shankar, Hongseok Namkoong, John Miller, Hannaneh Hajishirzi, Ali Farhadi, and Ludwig Schmidt.
\newblock Openclip, 2021.
\newblock If you use this software, please cite it as below.

\bibitem[Jaegle et~al.(2021)Jaegle, Gimeno, Brock, Zisserman, Vinyals, and Carreira]{jaegle2021perceivergeneralperceptioniterative}
Andrew Jaegle, Felix Gimeno, Andrew Brock, Andrew Zisserman, Oriol Vinyals, and Joao Carreira.
\newblock Perceiver: General perception with iterative attention, 2021.

\bibitem[Jiang et~al.(2023)Jiang, Sablayrolles, Mensch, Bamford, Chaplot, de~las Casas, Bressand, Lengyel, Lample, Saulnier, Lavaud, Lachaux, Stock, Scao, Lavril, Wang, Lacroix, and Sayed]{jiang2023mistral7b}
Albert~Q. Jiang, Alexandre Sablayrolles, Arthur Mensch, Chris Bamford, Devendra~Singh Chaplot, Diego de~las Casas, Florian Bressand, Gianna Lengyel, Guillaume Lample, Lucile Saulnier, Lélio~Renard Lavaud, Marie-Anne Lachaux, Pierre Stock, Teven~Le Scao, Thibaut Lavril, Thomas Wang, Timothée Lacroix, and William~El Sayed.
\newblock Mistral 7b, 2023.

\bibitem[Jie et~al.(2024)Jie, Tang, Ding, Deng, Han, and Wang]{jie2024memoryspacevisualpromptingefficient}
Shibo Jie, Yehui Tang, Ning Ding, Zhi-Hong Deng, Kai Han, and Yunhe Wang.
\newblock Memory-space visual prompting for efficient vision-language fine-tuning, 2024.

\bibitem[Kazemzadeh et~al.(2014)Kazemzadeh, Ordonez, Matten, and Berg]{kazemzadeh2014referitgame}
Sahar Kazemzadeh, Vicente Ordonez, Mark Matten, and Tamara Berg.
\newblock Referitgame: Referring to objects in photographs of natural scenes, 2014.

\bibitem[Krishna et~al.(2016)Krishna, Zhu, Groth, Johnson, Hata, Kravitz, Chen, Kalantidis, Li, Shamma, Bernstein, and Li]{krishna2016visualgenomeconnectinglanguage}
Ranjay Krishna, Yuke Zhu, Oliver Groth, Justin Johnson, Kenji Hata, Joshua Kravitz, Stephanie Chen, Yannis Kalantidis, Li-Jia Li, David~A. Shamma, Michael~S. Bernstein, and Fei-Fei Li.
\newblock Visual genome: Connecting language and vision using crowdsourced dense image annotations, 2016.

\bibitem[LAION(2023)]{OpenFlamingov2}
LAION.
\newblock Openflamingo v2: New models and enhanced training setup.
\newblock \url{https://laion.ai/blog/open-flamingo-v2/}, 2023.
\newblock Accessed: 2024-05-26.

\bibitem[Li et~al.(2023{\natexlab{a}})Li, Wang, Wang, Ge, Ge, and Shan]{li2023seedbenchbenchmarkingmultimodalllms}
Bohao Li, Rui Wang, Guangzhi Wang, Yuying Ge, Yixiao Ge, and Ying Shan.
\newblock Seed-bench: Benchmarking multimodal llms with generative comprehension, 2023{\natexlab{a}}.

\bibitem[Li et~al.(2023{\natexlab{b}})Li, Zhang, Chen, Wang, Yang, and Liu]{li2023ottermultimodalmodelincontext}
Bo Li, Yuanhan Zhang, Liangyu Chen, Jinghao Wang, Jingkang Yang, and Ziwei Liu.
\newblock Otter: A multi-modal model with in-context instruction tuning, 2023{\natexlab{b}}.

\bibitem[Li et~al.(2023{\natexlab{c}})Li, Li, Savarese, and Hoi]{li2023blip2bootstrappinglanguageimagepretraining}
Junnan Li, Dongxu Li, Silvio Savarese, and Steven Hoi.
\newblock Blip-2: Bootstrapping language-image pre-training with frozen image encoders and large language models, 2023{\natexlab{c}}.

\bibitem[Li et~al.(2023{\natexlab{d}})Li, Du, Zhou, Wang, Zhao, and Wen]{li2023evaluatingobjecthallucinationlarge}
Yifan Li, Yifan Du, Kun Zhou, Jinpeng Wang, Wayne~Xin Zhao, and Ji-Rong Wen.
\newblock Evaluating object hallucination in large vision-language models, 2023{\natexlab{d}}.

\bibitem[Li et~al.(2023{\natexlab{e}})Li, Wang, and Jia]{li2023llamavidimageworth2}
Yanwei Li, Chengyao Wang, and Jiaya Jia.
\newblock Llama-vid: An image is worth 2 tokens in large language models, 2023{\natexlab{e}}.

\bibitem[Lin et~al.(2023)Lin, Liu, Zhang, Gao, Qiu, Xiao, Qiu, Lin, Shao, Chen, Han, Huang, Zhang, He, Li, and Qiao]{lin2023sphinxjointmixingweights}
Ziyi Lin, Chris Liu, Renrui Zhang, Peng Gao, Longtian Qiu, Han Xiao, Han Qiu, Chen Lin, Wenqi Shao, Keqin Chen, Jiaming Han, Siyuan Huang, Yichi Zhang, Xuming He, Hongsheng Li, and Yu Qiao.
\newblock Sphinx: The joint mixing of weights, tasks, and visual embeddings for multi-modal large language models, 2023.

\bibitem[Liu et~al.(2023)Liu, Li, Wu, and Lee]{liu2023visualinstructiontuning}
Haotian Liu, Chunyuan Li, Qingyang Wu, and Yong~Jae Lee.
\newblock Visual instruction tuning, 2023.

\bibitem[Liu et~al.(2024{\natexlab{a}})Liu, Li, Li, and Lee]{liu2024improvedbaselinesvisualinstruction}
Haotian Liu, Chunyuan Li, Yuheng Li, and Yong~Jae Lee.
\newblock Improved baselines with visual instruction tuning, 2024{\natexlab{a}}.

\bibitem[Liu et~al.(2024{\natexlab{b}})Liu, Duan, Zhang, Li, Zhang, Zhao, Yuan, Wang, He, Liu, Chen, and Lin]{liu2024mmbenchmultimodalmodelallaround}
Yuan Liu, Haodong Duan, Yuanhan Zhang, Bo Li, Songyang Zhang, Wangbo Zhao, Yike Yuan, Jiaqi Wang, Conghui He, Ziwei Liu, Kai Chen, and Dahua Lin.
\newblock Mmbench: Is your multi-modal model an all-around player?, 2024{\natexlab{b}}.

\bibitem[Liu et~al.(2022)Liu, Mao, Wu, Feichtenhofer, Darrell, and Xie]{liu2022convnet2020s}
Zhuang Liu, Hanzi Mao, Chao-Yuan Wu, Christoph Feichtenhofer, Trevor Darrell, and Saining Xie.
\newblock A convnet for the 2020s, 2022.

\bibitem[Lu et~al.(2022)Lu, Mishra, Xia, Qiu, Chang, Zhu, Tafjord, Clark, and Kalyan]{lu2022learnexplainmultimodalreasoning}
Pan Lu, Swaroop Mishra, Tony Xia, Liang Qiu, Kai-Wei Chang, Song-Chun Zhu, Oyvind Tafjord, Peter Clark, and Ashwin Kalyan.
\newblock Learn to explain: Multimodal reasoning via thought chains for science question answering, 2022.

\bibitem[Ma et~al.(2024{\natexlab{a}})Ma, Xue, Wang, Zhou, Rao, Yan, Zhang, Wu, Shou, and Sun]{ma2024visualperceptionlargelanguage}
Feipeng Ma, Hongwei Xue, Guangting Wang, Yizhou Zhou, Fengyun Rao, Shilin Yan, Yueyi Zhang, Siying Wu, Mike~Zheng Shou, and Xiaoyan Sun.
\newblock Visual perception by large language model's weights, 2024{\natexlab{a}}.

\bibitem[Ma et~al.(2024{\natexlab{b}})Ma, Zhou, Li, He, Wu, Rao, Zhang, and Sun]{ma2024eemllmdataefficientcomputeefficientmultimodal}
Feipeng Ma, Yizhou Zhou, Hebei Li, Zilong He, Siying Wu, Fengyun Rao, Yueyi Zhang, and Xiaoyan Sun.
\newblock Ee-mllm: A data-efficient and compute-efficient multimodal large language model, 2024{\natexlab{b}}.

\bibitem[Mao et~al.(2016)Mao, Huang, Toshev, Camburu, Yuille, and Murphy]{mao2016generationcomprehensionunambiguousobject}
Junhua Mao, Jonathan Huang, Alexander Toshev, Oana Camburu, Alan Yuille, and Kevin Murphy.
\newblock Generation and comprehension of unambiguous object descriptions, 2016.

\bibitem[Marino et~al.(2019)Marino, Rastegari, Farhadi, and Mottaghi]{marino2019okvqavisualquestionanswering}
Kenneth Marino, Mohammad Rastegari, Ali Farhadi, and Roozbeh Mottaghi.
\newblock Ok-vqa: A visual question answering benchmark requiring external knowledge, 2019.

\bibitem[Meta(2024)]{llama3v}
Meta.
\newblock Introducing meta llama 3: The most capable openly available llm to date.
\newblock \url{https://ai.meta.com/blog/meta-llama-3/}, 2024.
\newblock Accessed: 2024-05-26.

\bibitem[Mishra et~al.(2019)Mishra, Shekhar, Singh, and Chakraborty]{mishra2019ocrvqa}
Anand Mishra, Shashank Shekhar, Ajeet~Kumar Singh, and Anirban Chakraborty.
\newblock Ocr-vqa: Visual question answering by reading text in images, 2019.

\bibitem[OpenAI(2023)]{gpt4v}
OpenAI.
\newblock Gpt-4v(ision) system card.
\newblock \url{https://cdn.openai.com/papers/GPTV_System_Card.pdf}, 2023.

\bibitem[OpenAI(2024)]{openai2024gpt4technicalreport}
OpenAI.
\newblock Gpt-4 technical report, 2024.

\bibitem[Radford et~al.(2021)Radford, Kim, Hallacy, Ramesh, Goh, Agarwal, Sastry, Askell, Mishkin, Clark, Krueger, and Sutskever]{radford2021learningtransferablevisualmodels}
Alec Radford, Jong~Wook Kim, Chris Hallacy, Aditya Ramesh, Gabriel Goh, Sandhini Agarwal, Girish Sastry, Amanda Askell, Pamela Mishkin, Jack Clark, Gretchen Krueger, and Ilya Sutskever.
\newblock Learning transferable visual models from natural language supervision, 2021.

\bibitem[Schuhmann et~al.(2022)Schuhmann, Beaumont, Vencu, Gordon, Wightman, Cherti, Coombes, Katta, Mullis, Wortsman, Schramowski, Kundurthy, Crowson, Schmidt, Kaczmarczyk, and Jitsev]{schuhmann2022laion5bopenlargescaledataset}
Christoph Schuhmann, Romain Beaumont, Richard Vencu, Cade Gordon, Ross Wightman, Mehdi Cherti, Theo Coombes, Aarush Katta, Clayton Mullis, Mitchell Wortsman, Patrick Schramowski, Srivatsa Kundurthy, Katherine Crowson, Ludwig Schmidt, Robert Kaczmarczyk, and Jenia Jitsev.
\newblock Laion-5b: An open large-scale dataset for training next generation image-text models, 2022.

\bibitem[Schwenk et~al.(2022)Schwenk, Khandelwal, Clark, Marino, and Mottaghi]{schwenk2022aokvqabenchmarkvisualquestion}
Dustin Schwenk, Apoorv Khandelwal, Christopher Clark, Kenneth Marino, and Roozbeh Mottaghi.
\newblock A-okvqa: A benchmark for visual question answering using world knowledge, 2022.

\bibitem[Shang et~al.(2024)Shang, Cai, Xu, Lee, and Yan]{shang2024llavaprumergeadaptivetokenreduction}
Yuzhang Shang, Mu Cai, Bingxin Xu, Yong~Jae Lee, and Yan Yan.
\newblock Llava-prumerge: Adaptive token reduction for efficient large multimodal models, 2024.

\bibitem[ShareGPT(2023)]{sharegpt2023}
ShareGPT.
\newblock \url{https://sharegpt.com/}, 2023.

\bibitem[Sidorov et~al.(2020)Sidorov, Hu, Rohrbach, and Singh]{sidorov2020textcapsdatasetimagecaptioning}
Oleksii Sidorov, Ronghang Hu, Marcus Rohrbach, and Amanpreet Singh.
\newblock Textcaps: a dataset for image captioning with reading comprehension, 2020.

\bibitem[Singh et~al.(2019)Singh, Natarajan, Shah, Jiang, Chen, Batra, Parikh, and Rohrbach]{singh2019vqamodelsread}
Amanpreet Singh, Vivek Natarajan, Meet Shah, Yu Jiang, Xinlei Chen, Dhruv Batra, Devi Parikh, and Marcus Rohrbach.
\newblock Towards vqa models that can read, 2019.

\bibitem[Tong et~al.(2024{\natexlab{a}})Tong, Brown, Wu, Woo, Middepogu, Akula, Yang, Yang, Iyer, Pan, Wang, Fergus, LeCun, and Xie]{tong2024cambrian1fullyopenvisioncentric}
Shengbang Tong, Ellis Brown, Penghao Wu, Sanghyun Woo, Manoj Middepogu, Sai~Charitha Akula, Jihan Yang, Shusheng Yang, Adithya Iyer, Xichen Pan, Ziteng Wang, Rob Fergus, Yann LeCun, and Saining Xie.
\newblock Cambrian-1: A fully open, vision-centric exploration of multimodal llms, 2024{\natexlab{a}}.

\bibitem[Tong et~al.(2024{\natexlab{b}})Tong, Liu, Zhai, Ma, LeCun, and Xie]{tong2024eyeswideshutexploring}
Shengbang Tong, Zhuang Liu, Yuexiang Zhai, Yi Ma, Yann LeCun, and Saining Xie.
\newblock Eyes wide shut? exploring the visual shortcomings of multimodal llms, 2024{\natexlab{b}}.

\bibitem[Touvron et~al.(2023{\natexlab{a}})Touvron, Lavril, Izacard, Martinet, Lachaux, Lacroix, Rozière, Goyal, Hambro, Azhar, Rodriguez, Joulin, Grave, and Lample]{touvron2023llamaopenefficientfoundation}
Hugo Touvron, Thibaut Lavril, Gautier Izacard, Xavier Martinet, Marie-Anne Lachaux, Timothée Lacroix, Baptiste Rozière, Naman Goyal, Eric Hambro, Faisal Azhar, Aurelien Rodriguez, Armand Joulin, Edouard Grave, and Guillaume Lample.
\newblock Llama: Open and efficient foundation language models, 2023{\natexlab{a}}.

\bibitem[Touvron et~al.(2023{\natexlab{b}})Touvron, Martin, Stone, Albert, Almahairi, Babaei, Bashlykov, Batra, Bhargava, Bhosale, Bikel, Blecher, Ferrer, Chen, Cucurull, Esiobu, Fernandes, Fu, Fu, Fuller, Gao, Goswami, Goyal, Hartshorn, Hosseini, Hou, Inan, Kardas, Kerkez, Khabsa, Kloumann, Korenev, Koura, Lachaux, Lavril, Lee, Liskovich, Lu, Mao, Martinet, Mihaylov, Mishra, Molybog, Nie, Poulton, Reizenstein, Rungta, Saladi, Schelten, Silva, Smith, Subramanian, Tan, Tang, Taylor, Williams, Kuan, Xu, Yan, Zarov, Zhang, Fan, Kambadur, Narang, Rodriguez, Stojnic, Edunov, and Scialom]{touvron2023llama2openfoundation}
Hugo Touvron, Louis Martin, Kevin Stone, Peter Albert, Amjad Almahairi, Yasmine Babaei, Nikolay Bashlykov, Soumya Batra, Prajjwal Bhargava, Shruti Bhosale, Dan Bikel, Lukas Blecher, Cristian~Canton Ferrer, Moya Chen, Guillem Cucurull, David Esiobu, Jude Fernandes, Jeremy Fu, Wenyin Fu, Brian Fuller, Cynthia Gao, Vedanuj Goswami, Naman Goyal, Anthony Hartshorn, Saghar Hosseini, Rui Hou, Hakan Inan, Marcin Kardas, Viktor Kerkez, Madian Khabsa, Isabel Kloumann, Artem Korenev, Punit~Singh Koura, Marie-Anne Lachaux, Thibaut Lavril, Jenya Lee, Diana Liskovich, Yinghai Lu, Yuning Mao, Xavier Martinet, Todor Mihaylov, Pushkar Mishra, Igor Molybog, Yixin Nie, Andrew Poulton, Jeremy Reizenstein, Rashi Rungta, Kalyan Saladi, Alan Schelten, Ruan Silva, Eric~Michael Smith, Ranjan Subramanian, Xiaoqing~Ellen Tan, Binh Tang, Ross Taylor, Adina Williams, Jian~Xiang Kuan, Puxin Xu, Zheng Yan, Iliyan Zarov, Yuchen Zhang, Angela Fan, Melanie Kambadur, Sharan Narang, Aurelien Rodriguez, Robert Stojnic, Sergey Edunov, and Thomas
  Scialom.
\newblock Llama 2: Open foundation and fine-tuned chat models, 2023{\natexlab{b}}.

\bibitem[Ye et~al.(2024)Ye, Gan, Huang, Ge, Shan, and Tang]{ye2024vocollamavisioncompressionlarge}
Xubing Ye, Yukang Gan, Xiaoke Huang, Yixiao Ge, Ying Shan, and Yansong Tang.
\newblock Voco-llama: Towards vision compression with large language models, 2024.

\bibitem[Yue et~al.(2024)Yue, Ni, Zhang, Zheng, Liu, Zhang, Stevens, Jiang, Ren, Sun, Wei, Yu, Yuan, Sun, Yin, Zheng, Yang, Liu, Huang, Sun, Su, and Chen]{yue2024mmmumassivemultidisciplinemultimodal}
Xiang Yue, Yuansheng Ni, Kai Zhang, Tianyu Zheng, Ruoqi Liu, Ge Zhang, Samuel Stevens, Dongfu Jiang, Weiming Ren, Yuxuan Sun, Cong Wei, Botao Yu, Ruibin Yuan, Renliang Sun, Ming Yin, Boyuan Zheng, Zhenzhu Yang, Yibo Liu, Wenhao Huang, Huan Sun, Yu Su, and Wenhu Chen.
\newblock Mmmu: A massive multi-discipline multimodal understanding and reasoning benchmark for expert agi, 2024.

\bibitem[Zhai et~al.(2023)Zhai, Mustafa, Kolesnikov, and Beyer]{zhai2023sigmoidlosslanguageimage}
Xiaohua Zhai, Basil Mustafa, Alexander Kolesnikov, and Lucas Beyer.
\newblock Sigmoid loss for language image pre-training, 2023.

\bibitem[Zhu et~al.(2023)Zhu, Chen, Shen, Li, and Elhoseiny]{zhu2023minigpt4enhancingvisionlanguageunderstanding}
Deyao Zhu, Jun Chen, Xiaoqian Shen, Xiang Li, and Mohamed Elhoseiny.
\newblock Minigpt-4: Enhancing vision-language understanding with advanced large language models, 2023.

\end{thebibliography}
